\documentclass[10pt,twocolumn,letterpaper]{article}

\usepackage{cvpr}

\usepackage{times}
\usepackage{relsize}
\usepackage[T1]{fontenc}
\usepackage[scaled=0.90]{inconsolata}
\usepackage[latin1]{inputenc}
\usepackage[english]{babel}

\usepackage{epsfig}
\usepackage{graphicx}
\usepackage{wrapfig}
\usepackage[belowskip=0pt,aboveskip=0pt,font=small]{caption}
\usepackage[belowskip=0pt,aboveskip=0pt,font=small]{subcaption}
\usepackage{longtable}

\usepackage{amsmath, amsthm, amssymb, mathabx, amstext}
\usepackage{textcomp}
\usepackage{stmaryrd}
\SetSymbolFont{stmry}{bold}{U}{stmry}{m}{n}
\usepackage{upgreek}
\usepackage{bm}
\usepackage{cases}
\usepackage{mathtools}
\usepackage{arydshln}
\usepackage{multirow}

\usepackage{color}
\usepackage[dvipsnames]{xcolor}
\definecolor{JalapenoRed}{RGB}{183,21,64}
\definecolor{Belize}{RGB}{41,128,185}
\definecolor{Amour}{RGB}{238,82,83}

\usepackage{cite}

\usepackage[pagebackref=true,breaklinks=true,letterpaper=true,colorlinks,bookmarks=false,allcolors=JalapenoRed]{hyperref}
\usepackage{bibspacing}
\usepackage{algorithm}
\usepackage{algpseudocode}

\usepackage{multirow}
\usepackage{rotating}
\usepackage{booktabs}

\usepackage{enumitem}
\usepackage[olditem,oldenum]{paralist}

\usepackage{alltt}
\usepackage{listings}

\usepackage{mysymbols}

\usepackage{url}
\usepackage{xspace}
\usepackage{comment}
\usepackage{afterpage}
\usepackage{pdfpages}
\usepackage{framed}
\usepackage{fancybox}
\usepackage{cuted}
\usepackage{caption}

\usepackage{quoting}
\quotingsetup{vskip=0pt}
\usepackage{epigraph}

\cvprfinalcopy %

\def\cvprPaperID{6382} %

\ifcvprfinal\fi
\begin{document}

\title{Large-scale Pretraining for Visual Dialog: \\A Simple State-of-the-Art Baseline}

\author{
    Vishvak Murahari$^1$, \,\,
    Dhruv Batra$^{2,1}$, \,\,
    Devi Parikh$^{2,1}$, \,\,
    Abhishek Das$^{1}$ \vspace{3pt}\\
    $^1$Georgia Institute of Technology, $^2$Facebook AI Research\\
    {\tt\small $^1$\{vishvak.murahari, parikh, dbatra, abhshkdz\}@gatech.edu} \\
    \tt\normalsize \href{https://visualdialog.org}{visualdialog.org}
}

\maketitle

\begin{abstract}
    Prior work in visual dialog has focused on training deep neural models on VisDial~\cite{visdial} in isolation. Instead, we present an approach to leverage pretraining on related vision-language datasets before transferring to visual dialog. We adapt the recently proposed ViLBERT model~\cite{lu_neurips19} for multi-turn visually-grounded conversations. Our model is pretrained on the Conceptual Captions~\cite{sharma_acl18} and Visual Question Answering~\cite{antol_iccv15} datasets, and finetuned on VisDial. Our best single model outperforms prior published work (including model ensembles) by more than $1\%$ absolute on NDCG and MRR. \\
    Next, we find that additional finetuning using "dense" annotations in VisDial leads to even higher NDCG -- more than $10\%$ over our base model -- but hurts MRR -- more than $17\%$ below our base model! This highlights a trade-off between the two primary metrics -- NDCG and MRR -- which we find is due to dense annotations not correlating well with the original ground-truth answers to questions.
\end{abstract}

\vspace{-20pt}
\section{Introduction}

Recent years have seen incredible progress in Visual
Dialog~\cite{vries_cvpr17,visdial,strub_arxiv17,visdial_rl,lu_nips17,massiceti_cvpr18,wu_cvpr18,jain_cvpr18,kottur_eccv18,lee_iclr19,niu_cvpr19,zheng_cvpr19,schwartz_cvpr19,kang_emnlp19,gan_acl2019,kottur_naacl19,visdial_diversity,shekhar_naacl19,yang_arxiv19b,guo_cvpr19,qi_arxiv19,jiang_aaai20},
spurred in part by the initial efforts of Das~\etal~\cite{visdial} in developing
a concrete task definition -- given an image, dialog history consisting of a sequence of
question-answer pairs, and a follow-up question about the image, to predict a
free-form natural language answer to the question -- along with a large-scale
dataset and evaluation metrics. The state-of-the-art on the task has improved
by more than $20\%$ absolute (${\sim}54\% \rightarrow {\sim}74\%$ NDCG) and the
original task has since been extended to challenging domains,~\eg
video understanding~\cite{alamri_cvpr19}, navigation assistants~\cite{vries_arxiv18,nguyen_emnlp19,thomason_corl19}.

\begin{figure*}[t]
    \includegraphics[width=\textwidth]{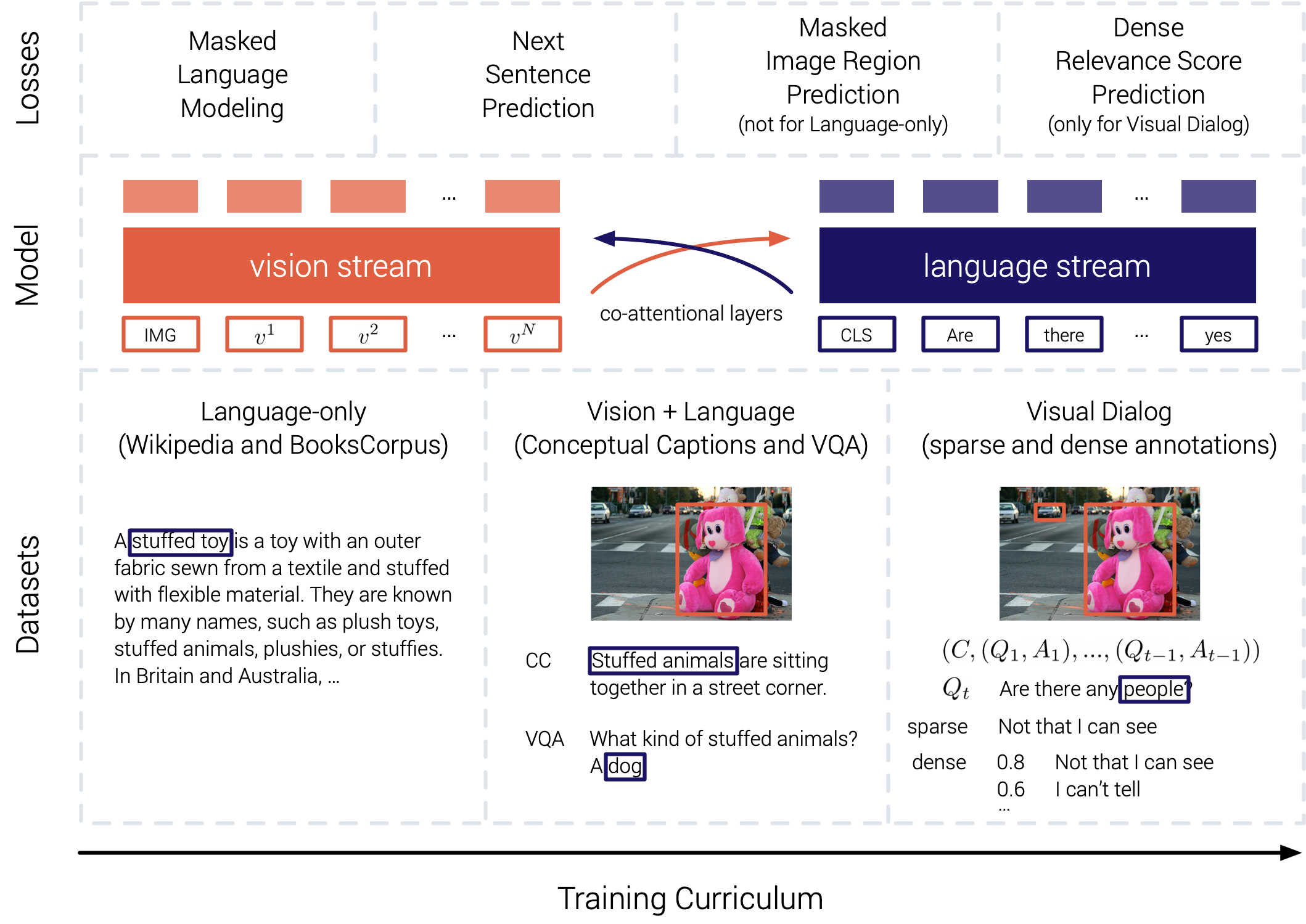}
    \vspace{2pt}
    \caption{First, the language stream of our model is pretrained on English Wikipedia and the BooksCorpus \cite{zhu_iccv15} datasets with the masked language modeling (MLM) and next sentence prediction (NSP) losses. Next, the entire model is trained on the Conceptual Captions \cite{sharma_acl18} and VQA \cite{antol_iccv15} datasets with the masked image region (MIR), MLM and NSP losses. Finally, we finetune the model on sparse annotations from VisDial \cite{visdial} with the MIR, MLM and NSP losses, and optionally finetune on dense annotations}
\vspace{-10pt}
\end{figure*}
While this is promising, much of this progress has happened in isolation,
wherein sophisticated neural architectures are trained and benchmarked solely
on the VisDial dataset. This is limiting -- since there is a significant amount
of shared abstraction and visual grounding in related tasks in vision and
language (\eg captioning, visual question answering) that can benefit Visual Dialog --
and wasteful -- since it is expensive and dissatisfying to have to collect a
large-scale dataset for every new task. In this work, we explore an
approach to pretrain our model on other related vision and
language datasets and then transfer to Visual Dialog.

Our work is inspired by prior work in transfer learning in
computer vision and natural language processing where large
models~\cite{he_cvpr16,simonyan_iclr15,krizhevsky_nips12,radford_2018,vaswani_nips17,devlin_naacl19,liu_arxiv19,lan_arxiv19,yang_arxiv19,raffel_arxiv19,zhang_arxiv19}
are pretrained on large datasets~\cite{russakovsky_ijcv15,krishna_ijcv17,zhu_iccv15}
with simple self-supervised objectives
to learn powerful representations that are then transferred to downstream tasks,
leading to state-of-the-art results on a variety
of benchmarks~\cite{russakovsky_ijcv15,wang_arxiv18}.
Recent work has extended this to vision and language
tasks~\cite{lu_neurips19,li_arxiv19,tan_arxiv19,chen_arxiv19,li_arxiv19b,su_arxiv19,sun_arxiv19},
leading to compelling results in Visual Question Answering\cite{antol_iccv15},
Commonsense Reasoning~\cite{zellers_cvpr19}, Natural Language Visual Reasoning~\cite{suhr_acl19},
Entailment~\cite{xie_arxiv19}, Image-Text Retrieval~\cite{young_tacl14,lee_eccv18},
Referring Expressions~\cite{kazemzadeh_emnlp14}, and Vision-Language Navigation~\cite{hao_cvpr20}.

In this work, we adapt ViLBERT~\cite{lu_neurips19} to Visual Dialog.
ViLBERT uses two Transformer-based\cite{vaswani_nips17} encoders, one
for each of the two modalities -- language and vision -- and interaction between the
two modalities is enabled by co-attention layers~\ie attention over inputs from
one modality conditioned on inputs from the other.
Note that adapting ViLBERT to Visual Dialog is not trivial.
The Visual Dialog dataset has image-grounded
conversation sequences that are up to $10$ rounds long. These are significantly
longer than captions (which are $\leq 2$ sentences) from the Conceptual Captions
dataset~\cite{sharma_acl18} or question-answer pairs from VQA~\cite{antol_iccv15}
used to pretrain ViLBERT, and thus requires
a different input representation and careful reconsideration of the
masked language modeling and next sentence prediction objectives
used to train BERT~\cite{devlin_naacl19} and ViLBERT~\cite{lu_neurips19}.

This adapted model outperforms prior published work by $> 1\%$ absolute
and achieves state-of-the-art on Visual Dialog. Next, we carefully analyse
our model and find that additional
finetuning on `dense' annotations\footnote{publicly available on
\href{https://visualdialog.org/data}{{\tt visualdialog.org/data}}.}~\ie
relevance scores for all $100$
answer options corresponding to each question on a subset of the training
set, highlights an interesting trade-off -- the model gets to
${\sim}74.5\%$ NDCG (outperforming the
2019 VisDial Challenge winner), but an MRR of ${\sim}52\%$
(${\sim}17\%$ below our base model!).
We find this happens because dense annotations in
VisDial do not correlate well with the ground-truth answers to
questions, often rewarding the model for generic, uncertain responses. \\

\vspace{-10pt}
Concretely, our contributions are as follows:
\begin{itemize}
    \item We introduce an adaptation of the ViLBERT~\cite{lu_neurips19} model
      for Visual Dialog, thus making use of the large-scale Conceptual Captions~\cite{sharma_acl18}
      and Visual Question Answering (VQA)~\cite{antol_iccv15} datasets for pretraining
      and learning powerful visually-grounded representations
      before finetuning on VisDial~\cite{visdial}. Since captioning and VQA
      differ significantly from Visual Dialog in input size ($\leq 2$
      sentence descriptions~\vs $\leq 10$ question-answer rounds), this requires
      rethinking the input representation to learn additional segment
      embeddings representing questions-answer pairs. Our adapted model improves
      over prior published work by $> 1\%$ and sets a new state-of-the-art.
    \item We next finetune our model on dense annotations~\ie
    relevance scores for all $100$ answer options corresponding to each question
    on a subset of the training set, leading to even higher NDCG -- more than
    $10\%$ over our base model -- but hurting MRR -- more than $17\%$ below our
    base model! This highlights a stark trade-off between the two primary metrics
    for this task -- NDCG and MRR. Through qualitative and
    quantitative results, we show that this happens because dense annotations do not
    correlate well with the original ground-truth answers, often rewarding the
    model for generic, uncertain responses.
    \item Our code is publicly
    available\footnote{\href{https://github.com/vmurahari3/visdial-bert}{{\tt https://github.com/vmurahari3/visdial-bert}}}
    to encourage further work in large-scale transfer learning for VisDial.
\end{itemize}
\section{Related Work}
\label{sec:related}

Our work is related to prior work in
visual dialog~\cite{vries_cvpr17,visdial,strub_arxiv17,visdial_rl,lu_nips17,massiceti_cvpr18,wu_cvpr18,jain_cvpr18,kottur_eccv18,lee_iclr19,niu_cvpr19,zheng_cvpr19,schwartz_cvpr19,kottur_naacl19,visdial_diversity,shekhar_naacl19,kang_emnlp19,gan_acl2019,yang_arxiv19b,guo_cvpr19,qi_arxiv19,jiang_aaai20},
and self-supervised pretraining and transfer learning in computer
vision and language~\cite{he_cvpr16,simonyan_iclr15,krizhevsky_nips12,radford_2018,vaswani_nips17,devlin_naacl19,liu_arxiv19,lan_arxiv19,yang_arxiv19,raffel_arxiv19,zhang_arxiv19}.

\textbf{Visual Dialog}.
Das~\etal~\cite{visdial} and de Vries~\etal~\cite{vries_cvpr17} introduced the
task of Visual Dialog -- given an image, dialog history consisting of a sequence
of question-answer pairs, and a follow-up question, predict a free-form natural
language answer to the question -- along with a dataset, evaluation metrics,
and baseline models. Follow-up works on visual dialog have explored the use of
deep reinforcement learning~\cite{strub_arxiv17,visdial_rl,visdial_diversity},
knowledge transfer from discriminative to generative decoders~\cite{lu_nips17},
conditional variational autoencoders~\cite{massiceti_cvpr18}, generative adversarial
networks~\cite{wu_cvpr18}, attention mechanisms for visual coreference
resolution~\cite{kottur_eccv18,niu_cvpr19}, and modeling the questioner's theory
of mind~\cite{lee_iclr19}. Crucially, all of these works train and evaluate
on the VisDial dataset \emph{in isolation}, without leveraging related visual
grounding signals from other large-scale datasets in vision and language.
We devise a unified model that can be pretrained on the Conceptual
Captions~\cite{sharma_acl18} and VQA~\cite{antol_iccv15}
datasets, and then transferred and finetuned on VisDial.

\textbf{Self-Supervised Learning in Vision and Language}.
Building on the success of transfer learning in natural language
understanding~\cite{radford_2018,vaswani_nips17,devlin_naacl19,liu_arxiv19,lan_arxiv19,yang_arxiv19,raffel_arxiv19,zhang_arxiv19}
leading to state-of-the-art results on a broad set of benchmarks~\cite{russakovsky_ijcv15,wang_arxiv18},
recent work has extended this to vision and language tasks~\cite{lu_neurips19,li_arxiv19,tan_arxiv19,chen_arxiv19,li_arxiv19b,su_arxiv19,sun_arxiv19}.
These works pretrain single~\cite{li_arxiv19,li_arxiv19b,su_arxiv19} or two~\cite{lu_neurips19,tan_arxiv19}-stream
Transformer~\cite{vaswani_nips17}-based models with self-supervised objectives, such as next-sentence prediction
and masked language/image modeling, on large-scale image-text datasets and have
led to compelling results in Visual Question Answering\cite{antol_iccv15},
Commonsense Reasoning~\cite{zellers_cvpr19}, Natural Language Visual Reasoning~\cite{suhr_acl19},
Entailment~\cite{xie_arxiv19}, Image-Text Retrieval~\cite{young_tacl14,lee_eccv18},
and Referring Expressions~\cite{kazemzadeh_emnlp14}, and Vision-Language Navigation~\cite{hao_cvpr20}.

\section{Adapting ViLBERT~\cite{lu_neurips19} for Visual Dialog}
\label{sec:approach}

Lu~\etal~\cite{lu_neurips19} introduced ViLBERT\footnote{along with code released at
\href{https://github.com/jiasenlu/ViLBERT\_beta}{{\tt github.com/jiasenlu/ViLBERT\_beta}}.},
which extended BERT~\cite{devlin_naacl19} to a two-stream multi-modal architecture for
jointly modeling visual and linguistic inputs. Interaction between the two
modalities was enabled through co-attention layers, \ie attending to
one modality conditioned on the other -- attention over language
conditioned on visual input, and attention over image regions conditioned on
linguistic input. This was operationalized as swapping the key
and value matrices between the visual and linguistic Transformer~\cite{vaswani_nips17}
blocks.
We next discuss our changes to adapt it for Visual Dialog followed by our
training pipeline.

\textbf{Input Representation}.
Recall that the model gets image $I$, dialog history (including image caption $C$)
$H = (C, (Q_1, A_1), ..., (Q_{t-1}, A_{t-1}))$, question $Q_t$, and a list of
$100$ answer options $A_t = \{A_t^{(1)}, A_t^{(2)}, ..., A_t^{(100)}\}$ as input,
and is asked to return a sorting of $A_t$.
We concatenate the $t$ rounds of dialog history and follow-up question $Q_t$,
with each question and answer separated by a {\tt <SEP>} token.
Similar to Wolf~\etal~\cite{wolf_arxiv19}, we use different segment embeddings
for questions and answers to help the model distinguish between the two
and understand question and answer boundaries in the input.
Captions and answers share the same segment embeddings.
To represent the image, we follow~\cite{lu_neurips19,anderson_arxiv17}
and extract object bounding boxes and their visual features for top-$36$ detected
objects in the image from a Faster R-CNN~\cite{ren_nips15b}
(with a ResNet-101~\cite{he_cvpr16} backbone) object detection network
pretrained on the Visual Genome dataset~\cite{krishna_ijcv17}.
The feature vector for each detected object is computed as mean-pooled
convolutional features from the regions of that object.
A $5$-d feature vector, consisting of normalized top-left and bottom-right
object coordinates, and the fraction of image area covered, is projected to
the same dimensions as the feature vector for the detected object, and added to it.
The beginning of this image region sequence (consisting of object detection
features) is demarcated by an {\tt IMG} token with mean-pooled features
from the entire image.

\subsection{Pretraining on Conceptual Captions~\cite{sharma_acl18}}
\label{sec:cctraining}
Following~\cite{lu_neurips19}, we pretrain the model on the
Conceptual Captions (CC) dataset, which is a large corpus (with ${\sim}3$M samples) of aligned
image-caption pairs. During pretraining, we optimize the sum of the masked
language modeling (MLM) loss~\cite{devlin_naacl19} and the masked image region (MIR)
loss. To compute the MLM loss, we mask around $15\%$ of the tokens in the input sequence and train the model 
to predict these tokens given context. Similarly, to compute the MIR loss we zero out $15\%$ of the image features and train the model to predict the semantic category 
of the masked out object (out of $1601$ classes from Visual Genome~\cite{krishna_ijcv17,anderson_arxiv17}).

\subsection{Pretraining on VQA~\cite{antol_iccv15}}
\label{sec:vqatraining}
The VQA dataset is quite related to Visual Dialog in that it can be interpreted
as independent visually-grounded question-answer pairs with no dialog history,
and thus is a natural choice for further pretraining prior to finetuning on VisDial.
Similar to Lu~\etal\cite{lu_neurips19}, we pretrain on VQA
by learning a small decoder -- a two-layer MLP -- on top of the element-wise
product between the image and text representations to predict a distribution
over $3129$ answers.

\subsection{Finetuning on Visual Dialog~\cite{visdial}}

To finetune on Visual Dialog, we use the MLM, NSP and the MIR losses. For MLM, we mask $10\%$ of the tokens in the dialog
sequence. For the MIR loss, similar to pretraining, we mask $15\%$ of the image features. Note that the discriminative task in visual dialog is to identify
the ground-truth answer from a list of $100$ answer options consisting of popular,
nearest neighbors, and random answers from the dataset. We achieve this through
the NSP loss.
The NSP head is trained to predict $1$ when the ground-truth answer is appended
to the input sequence, and $0$ when a negative answer sampled from the remaining
answer options is appended to it. Each image in VisDial has $10$ rounds of dialog,
leading to $10$ sets of positive and negative samples for the NSP loss per
mini-batch. Since these are fairly correlated samples, we randomly sub-sample
$2$ out of these $20$ during training.
At test time, we use log-probabilities from the NSP head to rank the $100$
answer options at each round.

\subsection{Finetuning with Dense Annotations}
\label{sec:denseft}
The authors of~\cite{visdial} recently released dense annotations\footnote{publicly available on
\href{https://visualdialog.org/data}{{\tt visualdialog.org/data}}.}~\ie
relevance scores for all $100$ answer options from $A_t$ corresponding to the
question on a subset of the training set. These relevance scores range from $0$
to $1$ and are calculated as the ratio of number of human annotators who marked
a particular answer option as correct to the total number of human annotators ($=4$).
So $1$ means that the answer option was considered correct by $4$ human
annotators. In our final stage of training, we
utilize these dense annotations to finetune our model.
Concretely, we use the NSP head to predict likelihood scores $\hat \ell_t^{(i)}$
for each answer option $A_t^{(i)}$ at round $t$, normalize these to form a probability
distribution over the $100$ answers $\hat y_t = [ \hat y_t^{(1)}, ..., \hat y_t^{(100)} ]$,
and then compute a cross-entropy (CE) loss against the normalized ground-truth
relevance scores $y_t$, given by $-\sum_i y_t^{(i)} \log \hat y_t^{(i)}$.

\section{Experiments}
\label{sec:experiments}

To compare to previous research, we conduct experiments on VisDial v$1.0$~\cite{visdial}.
The dataset contains human-human dialogs on ${\sim}130k$ COCO~\cite{mscoco}-like images. We follow
the original splits and use ${\sim}120k$ for training, ${\sim}2k$ for validation,
and ${\sim}8k$ for testing. We next describe the various settings we experiment with.

\textbf{Evaluation Metrics}. We use metrics introduced in~\cite{visdial}.
Specifically, given the predicted ranking of $100$ answer options from a model
at each round, we compute retrieval metrics -- mean rank (MR) of the ground-truth
answer, mean reciprocal rank (MRR), and recall@$k$ ($k=\{1,5,10\}$). Additionally,
along with the release of dense annotations,~\ie relevance scores
$\in [0, 1]$ for all $100$ answer options, a new metric -- NDCG -- was
introduced. NDCG accounts for multiple correct answers in the option set and
penalizes low-ranked but correct answer options.

\subsection{Language-only}
\label{sec:lonly}
We begin with a `blind' setting, where given the dialog history and follow-up
question, and without access to the image, the model is tasked with predicting
the answer. As such, we do not use the ViLBERT formulation for these experiments,
and finetune the BERT model released in~\cite{devlin_naacl19} and pretrained on
BooksCorpus~\cite{zhu_iccv15} and English Wikipedia.
For the MLM loss, we mask $15\%$ of tokens and sub-sample $8$ out of 20 sequences
per mini-batch during training. We experiment with two variants -- training only
with NSP, and training with both NSP and NSP.
See~\tableref{tab:val_results} for language-only results (marked `L-only').
This experimental setting helps us put gains coming from switching to
Transformer~\cite{vaswani_nips17}-based architectures (and before the added
complexity of incorporating visual input) in perspective.

\textbf{Varying number of dialog rounds}.
We train ablations of our language-only model (with NSP and MLM losses)
where we vary the number of rounds in dialog history, starting from $0$, where the
input sequence only contains the follow-up question and answer, to $2$, $4$, and
$6$ and $10$ rounds of dialog history (\tableref{tab:round_results}).

\textbf{Zero-shot and `cheap' finetuning}.
\label{sec:notrain}
We report performance for ablations of our NSP+MLM model with no/minimal training in~\tableref{tab:zero}.
First, we do a zero-shot test where we initialize BERT with weights from
Wikipedia and BooksCorpus pretraining
and simply run inference on VisDial. Second, with the same initialization, we freeze all
layers and finetune only the MLM and NSP loss heads.

\subsection{Finetuning on VisDial}
\label{sec:visdialft2}
We finetune ViLBERT on VisDial with three different weight initializations --
1) from the best language-only weights (from~\secref{sec:lonly}) for the
language stream (visual stream and co-attention layers initialized randomly),
2) from a model pretrained on CC~\cite{sharma_acl18} (as described in~\secref{sec:cctraining}),
and 3) from a model pretrained on CC~\cite{sharma_acl18}+VQA~\cite{antol_iccv15}
(as described in~\secref{sec:vqatraining}). 1) helps us benchmark performance
if the model learns visual grounding solely from VisDial, 2) quantifies
effects of learning visual grounding additionally from CC, while 3) helps
us quantify improvements with additional exposure to visually-grounded
question-answering data. See~\tableref{tab:val_results} for results.

\subsection{Finetuning with Dense Annotations}
\label{sec:denseft2}
Finally, we finetune our best model from~\secref{sec:visdialft2} -- marked
`w/ CC+VQA' in~\tableref{tab:val_results} -- on dense annotations, as described
in~\secref{sec:denseft}. Note that computing the CE loss requires a
separate forward pass for each of the $100$ answer options, since dialog history,
question, answer are all concatenated together before passing as input. This is
memory-expensive, and so in practice, we sub-sample and only use $80$ options,
and use gradient accumulation to (artificially) construct a larger mini-batch.
Finetuning with the CE loss only leads to significant
improvements on NDCG but hurts other metrics (see~\tableref{tab:val_results}).
We discuss and analyse this  in more detail later. But to control for this `metric-overfitting',
we also train a variant with both the CE and NSP losses.

\section{Results}
\label{sec:results}

We list findings from all our experiments in~\secref{sec:experiments} below,
and list additional negative results in the appendix.

\begin{table}[ht!]
    \resizebox{\columnwidth}{!}{
        \begin{tabular}{@{}crrrrrrrr@{}}
            & \multicolumn{6}{c}{}  \\[0.01in]
          $\#$ history rounds & NDCG $\uparrow$ & MRR $\uparrow$ & R@$1$ $\uparrow$ & R@$5$ $\uparrow$ & R@$10$ $\uparrow$ & MR $\downarrow$ \\[0.05in]
            \toprule
	$0$ & $50.54$ & $54.29$ & $38.88$ & $72.67$ & $83.09$ & $5.90$ \\
	$2$ & $53.69$  & $61.31$ & $46.83$ & $78.96$ & $88.15$ & $4.51$ \\
	$4$ & $55.10$ & $62.83$ & $48.36$ & $80.61$ & $89.57$ & $4.19$  \\
	$6$ & $55.69$ & $63.73$ & $49.31$ & $81.13$ & $90.06$ & $4.04$  \\
	$10$ & $\mathbf{57.26}$ & $\mathbf{64.40}$ & $\mathbf{50.30}$ & $\mathbf{81.60}$ & $\mathbf{90.43}$ & $\mathbf{4.01}$ \\
            \bottomrule
        \end{tabular}}
    \vspace{5pt}
    \caption{{\small Performance of the NSP + MLM language-only model on VisDial v$1.0$ val
    as the number of dialog history rounds is varied}}
    \label{tab:round_results}
    \vspace{-5pt}
\end{table}

\begin{table}[ht!]
    \resizebox{\columnwidth}{!}{
        \begin{tabular}{@{}lrrrrrrrr@{}}
            & \multicolumn{6}{c}{}  \\[0.01in]
            Model & NDCG $\uparrow$ & MRR $\uparrow$ & R@$1$ $\uparrow$ & R@$5$ $\uparrow$ & R@$10$ $\uparrow$ & MR $\downarrow$   \\[0.05in]
            \toprule
                No training & $11.63$ & $6.88$ & $2.63$ & $7.17$ & $11.30$ & $46.90$  \\
                Loss heads only & $19.69$ & $9.81$ & $3.42$ &  $10.44$ & $18.85$ & $31.38$  \\
            \bottomrule
        \end{tabular}}
    \vspace{5pt}
    \caption{{\small Performance of the NSP + MLM language-only model on VisDial v$1.0$ val
    with no / minimal training (described in Sec.~\ref{sec:notrain})}}
    \label{tab:zero}
\end{table}

\begin{figure*}[ht]
    \centering
    \begin{subfigure}[t]{0.27\textwidth}
        \centering
        \includegraphics[width=\textwidth]{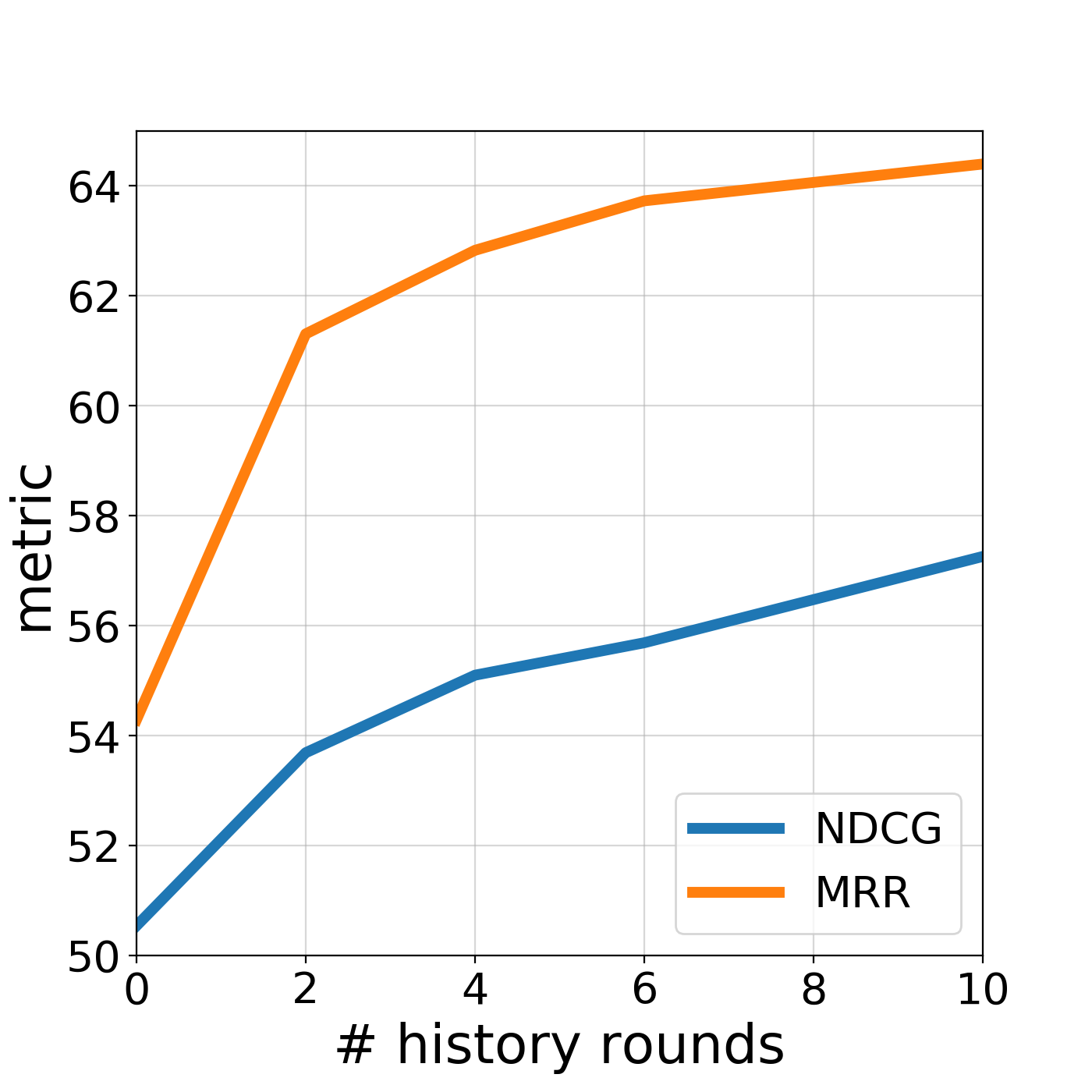}
        \caption{Change in metrics for varying number of rounds in dialog history for the language-only model trained with NSP + LM}
        \label{fig:dialog_history_viz}
    \end{subfigure}
    \quad
    \begin{subfigure}[t]{0.28\textwidth}
        \includegraphics[width=\textwidth]{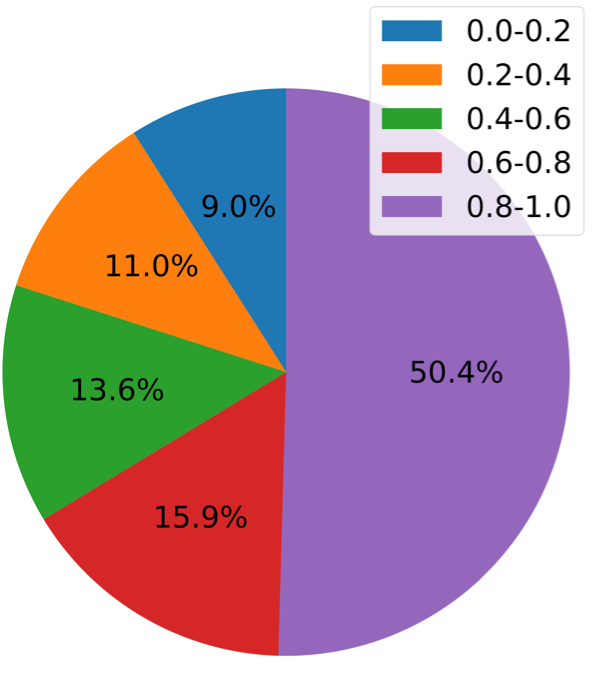}
        \caption{Distribution of dense annotation relevance scores for ground-truth
            answers in VisDial v$1.0$ val. ${\sim}50\%$ ground-truth answers
            have relevance scores $< 0.8$, and ${\sim}10\%$ have scores $< 0.2$}
        \label{fig:distribution_relevance_gt_answers}
    \end{subfigure}
    \quad
    \begin{subfigure}{0.38\textwidth}
        \centering
        \vspace{-30pt}
        \resizebox{\columnwidth}{!}{
            \begin{tabular}{@{}lrrrr@{}} \\[0.01in]
            Relevance Score & w/ CC + VQA & CE & CE + NSP \\[0.05in]
                \toprule
        $0.0-0.2$ & $6.47$ & $14.88$ & $10.79$  \\
        $0.2-0.4$ & $4.77$ & $11.11$  & $6.62$  \\
        $0.4-0.6$ & $4.02$ & $8.49$  & $4.86$    \\
        $0.6-0.8$ & $3.12$ & $6.63$  & $3.77$  \\
        $0.8-1.0$ & $1.95$ & $3.26$  & $2.21$  \\
                \bottomrule
            \end{tabular}}
        \vspace{10pt}
        \caption{Mean rank (lower is better) of the GT answers on VisDial v$1.0$ val split across model variants and ranges of relevance scores}
        \label{tab:relevance_distribution_gt_answer}
    \end{subfigure}
    \caption{}
    \label{fig:densedist}
    \vspace{-10pt}
\end{figure*}

\begin{itemize}[leftmargin=5pt]
    \item \textbf{Language-only performs well}.
        The language-only model gets to $57.26$ on NDCG and $64.40$ on MRR (\tableref{tab:val_results}). This
        is high and already competitive with several prior published works (see~\tableref{tab:teststd}).
    \item \textbf{Increasing dialog history rounds helps.}
        We report performance of the language-only model as a function of dialog
        history rounds in~\tableref{tab:round_results} and ~\figref{fig:dialog_history_viz}. Note that the change
        in performance from including $0$ to $4$ rounds of dialog history
        ($+4.56$ on NDCG, $+8.54$ on MRR) is much more than from $4$ to $10$
        dialog history rounds ($+2.16$ on NDCG, $+1.57$ on MRR). Thus, performance
        continues to go up with increasing dialog
        history rounds but starts to plateau with $\geq 4$ history rounds.
        We believe these improvements are largely indicative of the Transformer's
        ability to model long-term dependencies.
    \item \textbf{Zero-shot model performs poorly}. Running inference with
        the language-only model pretrained on BooksCorpus~\cite{zhu_iccv15} and Wikipedia
        without any finetuning on VisDial only gets to
        $11.63$ on NDCG and $6.88$ on MRR (\tableref{tab:zero}). Finetuning the loss heads with all other
        layers frozen leads to an improvement of ${\sim}8$ NDCG points over this.
        This low performance can be attributed to significantly longer sequences
        in VisDial than the model was pretrained with.
    \item \textbf{VQA initialization helps more than CC}. Finetuning ViLBERT
        on VisDial with weights initialized from VQA pretraining gets to
        $64.94$ on NDCG and $69.10$ on MRR, ${\sim}4$ points more than CC pretraining (\tableref{tab:val_results}).
        This is likely because images in VQA and VisDial are from COCO, as opposed to CC, and because the visual question answering task is more similar to VisDial than captioning.
    \item \textbf{Dense annotations boost NDCG, hurt MRR}.
        Finetuning  with the CE loss leads to $74.47$ on NDCG -- a ${\sim}10\%$
        improvement over the `w/ CC + VQA' base model -- but $50.74$ on MRR, a
        ${\sim}17\%$ decline below the base model (\tableref{tab:teststd}). This is a surprising
        finding! We carefully analyze this behavior
        in~\secref{sec:analysis}.
    \item \textbf{Ensembling does not improve performance.} We trained
        $3$ models initialized with different random seeds for each of the $3$
        variants (`w/ CC + VQA', `CE' and `CE + NSP') and aggregated results by
        averaging the normalized scores from the $3$ models. We did not observe
        any significant improvement.
\end{itemize}

\begin{table}[t!]
    \resizebox{\columnwidth}{!}{
        \begin{tabular}{@{}clrrrrrrrr@{}}
            & \multicolumn{6}{c}{}  \\[0.01in]
            & Model & NDCG $\uparrow$ & MRR $\uparrow$ & R@$1$ $\uparrow$ & R@$5$ $\uparrow$ & R@$10$ $\uparrow$ & MR $\downarrow$ \\[0.05in]
            \toprule
            \multirow{2}{*}{\rotatebox[origin=c]{90}{L-only} $\begin{dcases} \\ \\ \end{dcases}$}
            & NSP & $56.17$ & $63.37$ & $49.17$ & $80.62$ & $89.42$ & $4.21$ \\
            & NSP + MLM & $57.26$ & $64.40$ & $50.30$ & $81.60$ & $90.43$ & $4.01$ \\[0.1in]
            \multirow{2}{*}{\rotatebox[origin=c]{90}{$+$vision} $\begin{dcases} \\ \\ \\ \end{dcases}$}
            & w/ L-only & $62.64$ & $67.86$ & $54.54$ & $84.34$ & $92.36$ & $3.44$ \\
            & w/ CC~\cite{sharma_acl18} & $60.80$ & $67.13$ & $53.59$ & $84.39$ & $92.49$ & $3.44$ \\
           & w/ CC~\cite{sharma_acl18}+VQA~\cite{antol_iccv15} & $64.94$ & $\mathbf{69.10}$ & $\mathbf{55.88}$ & $\mathbf{85.50}$ & $\mathbf{93.29}$ & $\mathbf{3.25}$ \\[0.1in]
            \multirow{2}{*}{\rotatebox[origin=c]{90}{$+$dense} $\begin{dcases} \\ \\ \end{dcases}$}
                & CE & $\mathbf{75.24}$ & $52.22$ & $39.92$ & $65.05$ & $80.63$ & $6.17$ \\
                & CE + NSP & $69.24$ & $65.88$ & $53.41$ & $80.92$ & $90.18$ & $4.24$ \\[0.1in]
            \bottomrule
        \end{tabular}}
    \vspace{5pt}
    \caption{{\small Results on VisDial v$1.0$ val. $\uparrow$
    indicates higher is better}}
    \label{tab:val_results}
    \vspace{-10pt}
\end{table}
\begin{table}[ht!]
    \resizebox{\columnwidth}{!}{
        \vspace{-5pt}
        \begin{tabular}{@{}clrrrrrrrr@{}}
            & \multicolumn{6}{c}{}  \\[0.01in]
            & Model & NDCG $\uparrow$ & MRR $\uparrow$ & R@$1$ $\uparrow$ & R@$5$ $\uparrow$ & R@$10$ $\uparrow$ & MR $\downarrow$ \\[0.05in]
            \toprule
            \multirow{14}{*}{\rotatebox[origin=c]{90}{Published Results} $\begin{dcases} \\ \\ \\ \\ \\ \\ \\ \\ \\ \\ \\ \end{dcases}$}
            & GNN~\cite{zheng_cvpr19}                   & $52.82$ & $61.37$ & $47.33$ & $77.98$ & $87.83$ & $4.57$ \\
            & CorefNMN~\cite{kottur_eccv18}             & $54.70$ & $61.50$ & $47.55$ & $78.10$ & $88.80$ & $4.40$ \\
            & RvA~\cite{niu_cvpr19}                     & $55.59$ & $63.03$ & $49.03$ & $80.40$ & $89.83$ & $4.18$ \\
            & HACAN~\cite{yang_arxiv19b}                & $57.17$ & $64.22$ & $50.88$ & $80.63$ & $89.45$ & $4.20$ \\
            & NMN~\cite{kottur_eccv18}                  & $58.10$ & $58.80$ & $44.15$ & $76.88$ & $86.88$ & $4.81$ \\
            & DAN~\cite{kang_emnlp19}                   & $57.59$ & $63.20$ & $49.63$ & $79.75$ & $89.35$ & $4.30$ \\
            & DAN$^\dag$~\cite{kang_emnlp19}            & $59.36$ & $64.92$ & $51.28$ & $81.60$ & $90.88$ & $3.92$ \\
            & ReDAN~\cite{gan_acl2019}                  & $61.86$ & $53.13$ & $41.38$ & $66.07$ & $74.50$ & $8.91$ \\
            & ReDAN+$^\dag$~\cite{gan_acl2019}          & $64.47$ & $53.74$ & $42.45$ & $64.68$ & $75.68$ & $6.64$ \\
            & DualVD~\cite{jiang_aaai20}                & $56.32$ & $63.23$ & $49.25$ & $80.23$ & $89.70$ & $4.11$ \\
            & FGA~\cite{schwartz_cvpr19}                & $56.93$ & $66.22$ & $52.75$ & $82.92$ & $91.08$ & $3.81$ \\
            & DL-61~\cite{guo_cvpr19}                   & $57.32$ & $62.20$ & $47.90$ & $80.43$ & $89.95$ & $4.17$ \\
            & DL-61$^\dag$~\cite{guo_cvpr19}            & $57.88$ & $63.42$ & $49.30$ & $80.77$ & $90.68$ & $3.97$ \\
            & MReal - BDAI$^\star$~\cite{qi_arxiv19}    & $74.02$ & $52.62$ & $40.03$ & $68.85$ & $79.15$ & $6.76$ \\
            \toprule
            \multirow{10}{*}{\rotatebox[origin=c]{90}{Leaderboard Entries} $\begin{dcases} \\ \\ \\ \\ \\ \\ \\ \\ \end{dcases}$}
            & LF                                        & $45.31$ & $55.42$ & $40.95$ & $72.45$ & $82.83$ & $5.95$ \\
            & HRE                                       & $45.46$ & $54.16$ & $39.93$ & $70.45$ & $81.50$ & $6.41$ \\
            & MN                                        & $47.50$ & $55.49$ & $40.98$ & $72.30$ & $83.30$ & $5.92$ \\
            & MN-Att                                    & $49.58$ & $56.90$ & $42.43$ & $74.00$ & $84.35$ & $5.59$ \\
            & LF-Att                                    & $51.63$ & $60.41$ & $46.18$ & $77.80$ & $87.30$ & $4.75$ \\
            & MS ConvAI                                 & $55.35$ & $63.27$ & $49.53$ & $80.40$ & $89.60$ & $4.15$ \\
            & USTC-YTH                                  & $56.47$ & $61.44$ & $47.65$ & $78.13$ & $87.88$ & $4.65$ \\
            & UET-VNU                                   & $57.40$ & $59.50$ & $45.50$ & $76.33$ & $85.82$ & $5.34$ \\
            & square                                    & $60.16$ & $61.26$ & $47.15$ & $78.73$ & $88.48$ & $4.46$ \\
            & MS D365 AI                                & $64.47$ & $53.73$ & $42.45$ & $64.68$ & $75.68$ & $6.63$ \\
            \toprule
            \multirow{2}{*}{\rotatebox[origin=c]{90}{Ours} $\begin{dcases} \\ \\ \\ \end{dcases}$}
            & w/ CC~\cite{sharma_acl18}+VQA~\cite{antol_iccv15} & $63.87$ & $\mathbf{67.50}$ & $\mathbf{53.85}$ & $\mathbf{84.68}$ & $\mathbf{93.25}$  & $\mathbf{3.32}$ \\
            & CE                         & $\mathbf{74.47}$ & $50.74$ & $37.95$ & $64.13$ & $80.00$ & $6.28$ \\
            & CE $+$ NSP                 & $68.08$ & $63.92$ & $50.78$ & $79.53$ & $89.60$ & $4.28$ \\
        \end{tabular}}
    \vspace{5pt}
    \caption{{\small Summary of results on VisDial v$1.0$ test-std. $\uparrow$
    indicates higher is better. $\downarrow$ indicates lower is better. $\dag$
    denotes ensembles\\
    $\star$ denotes the winning team of the 2019 Visual Dialog Challenge.}}
    \label{tab:teststd}
    \vspace{-15pt}
\end{table}

We report results from the Visual Dialog evaluation
server\footnote{\href{https://evalai.cloudcv.org/web/challenges/challenge-page/161/leaderboard/483}{{\tt evalai.cloudcv.org/web/challenges/challenge-page/161/leaderboard/483}}}
for our best models -- `w/ CC + VQA', `CE' and `CE + NSP' --
on the unseen {\tt test-std} split in~\tableref{tab:teststd}. We compare against
prior published results and top entries from the leaderboard.
Our models outperform prior results and set a new state-of-the-art --
ViLBERT with CC + VQA pretraining on MRR, R@$k$, MR metrics, and further
finetuning with a CE loss on dense annotations on NDCG.
Finally, adding NSP loss along with CE (as in~\secref{sec:denseft2})
offers a balance between optimizing metrics that reward
both sparse (original ground-truth answers) and dense annotations.

\begin{figure*}[ht]
        \includegraphics[width=\textwidth]{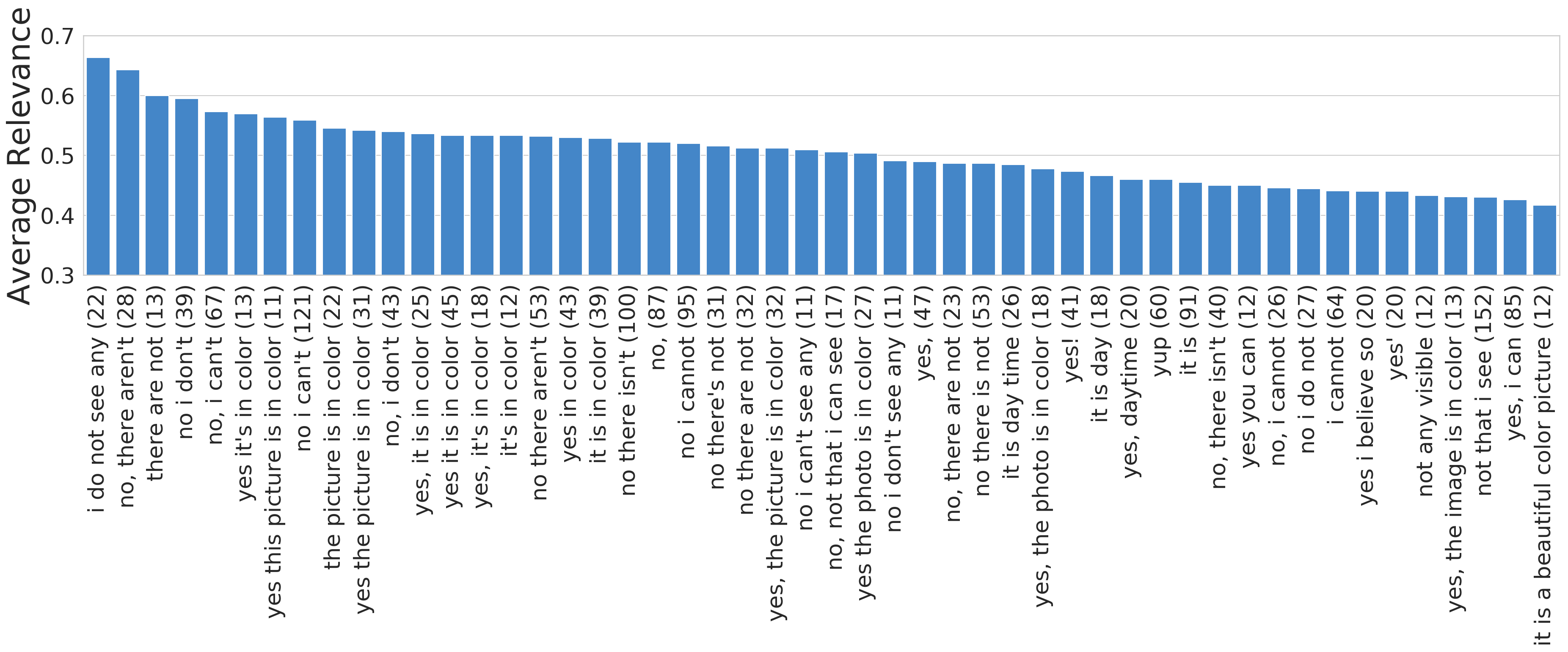}
        \caption{Mean relevance scores and counts for top-$50$ most-relevant answers
            from VisDial v$1.0$ val dense annotations. These contain several
            sets of paraphrases -- $\{$``yes it's in color'', ``yes this picture
            is in color'', ``the picture is in color'', ``yes the picture is in color'',
            ``yes, it is in color'', ``yes it is in color'', ``yes, it's in color'',
            ``yes in color''$\}$,~\etc and have a bias towards binary answers}
        \label{fig:all_val_top_50_options_avg_relevance}
        \vspace{-10pt}
\end{figure*}

\section{Analysis}
\label{sec:analysis}

\begin{figure*}[ht]
    \includegraphics[width=\textwidth]{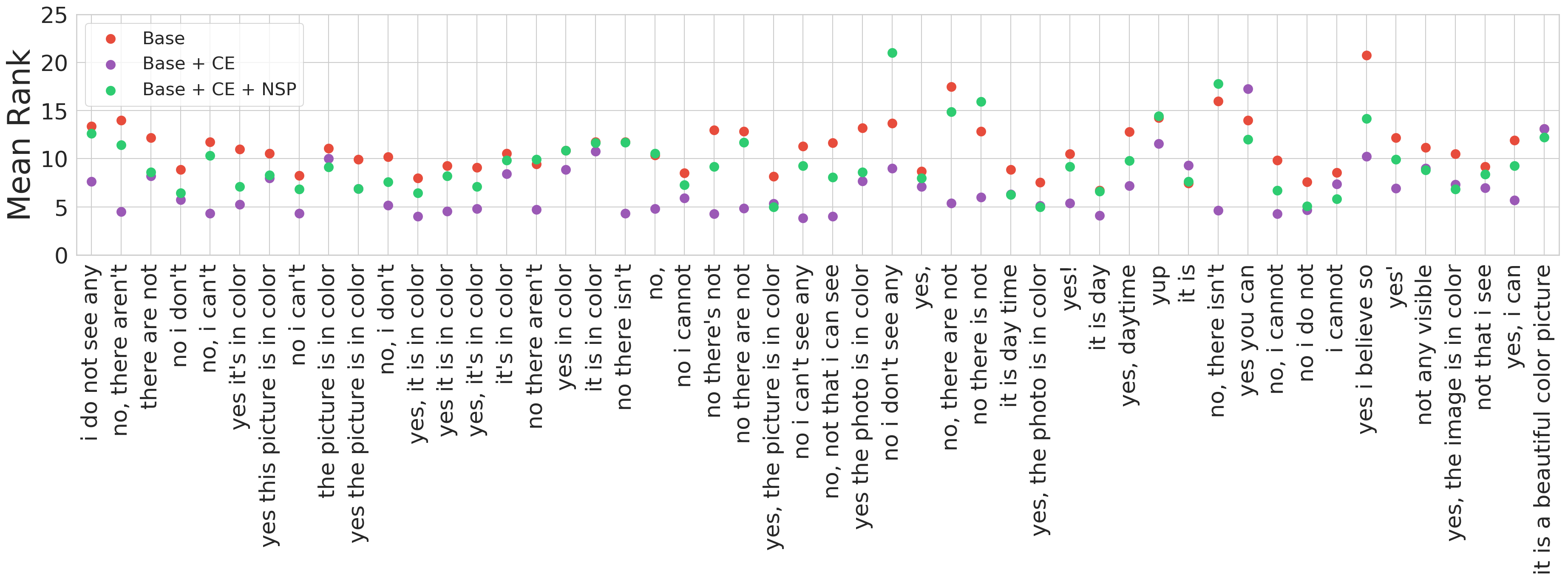}
    \caption{Predicted mean rank for each of the top-$50$ most relevant answers
        as per dense annotations (from~\figref{fig:all_val_top_50_options_avg_relevance})
        by three model variants -- ViLBERT w/ CC + VQA (called `Base'), CE, and CE + NSP.
        The CE model gets lower mean ranks for most answers in this set compared to Base.
        This leads to significantly higher NDCG, as reported in~\tableref{tab:val_results}
        and~\tableref{tab:teststd}, but low MRR, since these relevant answers as per
        dense annotations do not correlate well with the set of original ground-truth answers,
        as shown in~\figref{fig:distribution_relevance_gt_answers}}
    \label{fig:all_val_top_50_options_mean_rank}
    \vspace{-12pt}
\end{figure*}

As described in~\secref{sec:results}, finetuning on dense annotations leads to
a significant increase in NDCG, but hurts the other $5$ metrics -- MRR, R@$1$,
R@$5$, R@$10$ and MR -- which depend on the original sparse annotations in
VisDial~\ie follow-up answers provided in human-human dialog.

We begin by visualizing the distribution of dense relevance scores for these
sparse ground-truth (GT) answers in~\figref{fig:distribution_relevance_gt_answers}
and observe that ${\sim}50\%$ GT answers have relevance $\leq 0.8$, and
${\sim}30\%$ have relevance $\leq 0.6$. Thus, there is some degree of
misalignment between dense and sparse annotations -- answers originally provided
during human-human dialog in VisDial were not always judged to be relevant by all
humans during the post-hoc dense annotation phase.

\begin{figure*}[ht]
    \begin{subfigure}[t]{\textwidth}
        \includegraphics[width=\textwidth]{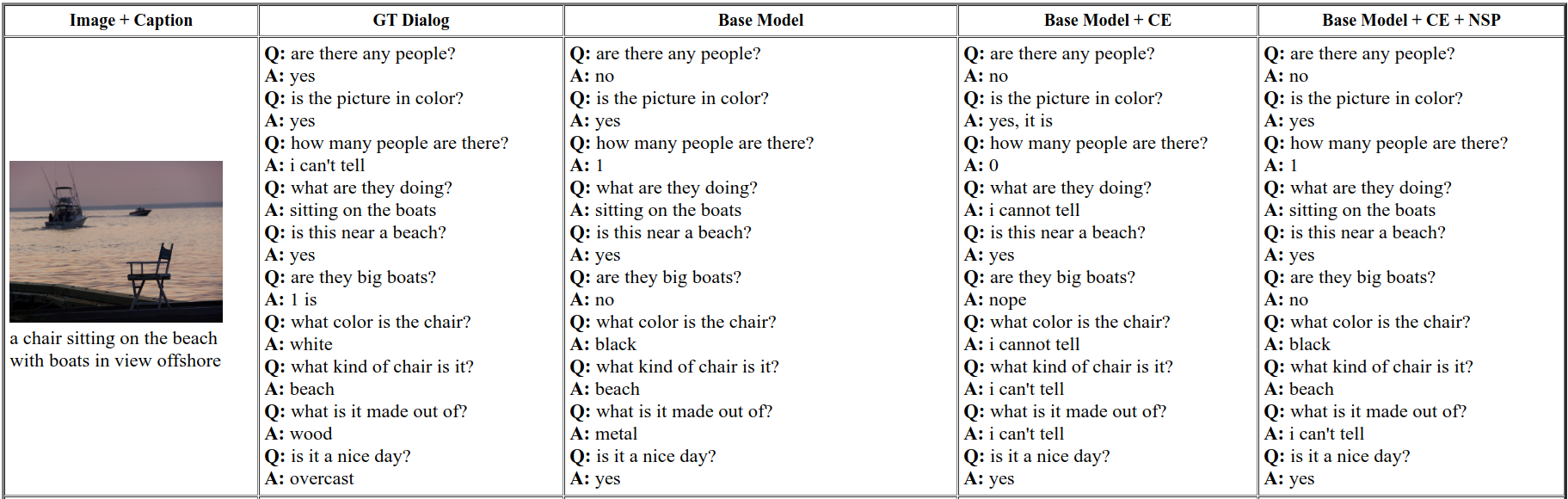}
    \end{subfigure}
    \begin{subfigure}[t]{\textwidth}
        \includegraphics[width=\textwidth]{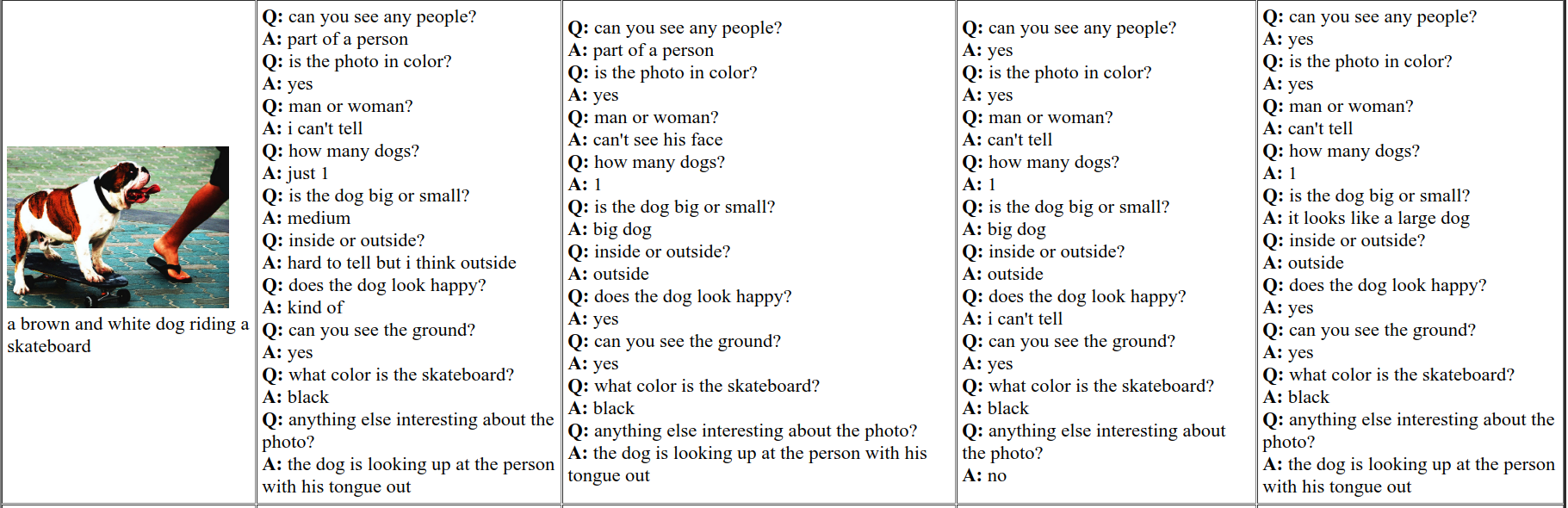}
    \end{subfigure}
    \begin{subfigure}[t]{\textwidth}
        \includegraphics[width=\textwidth]{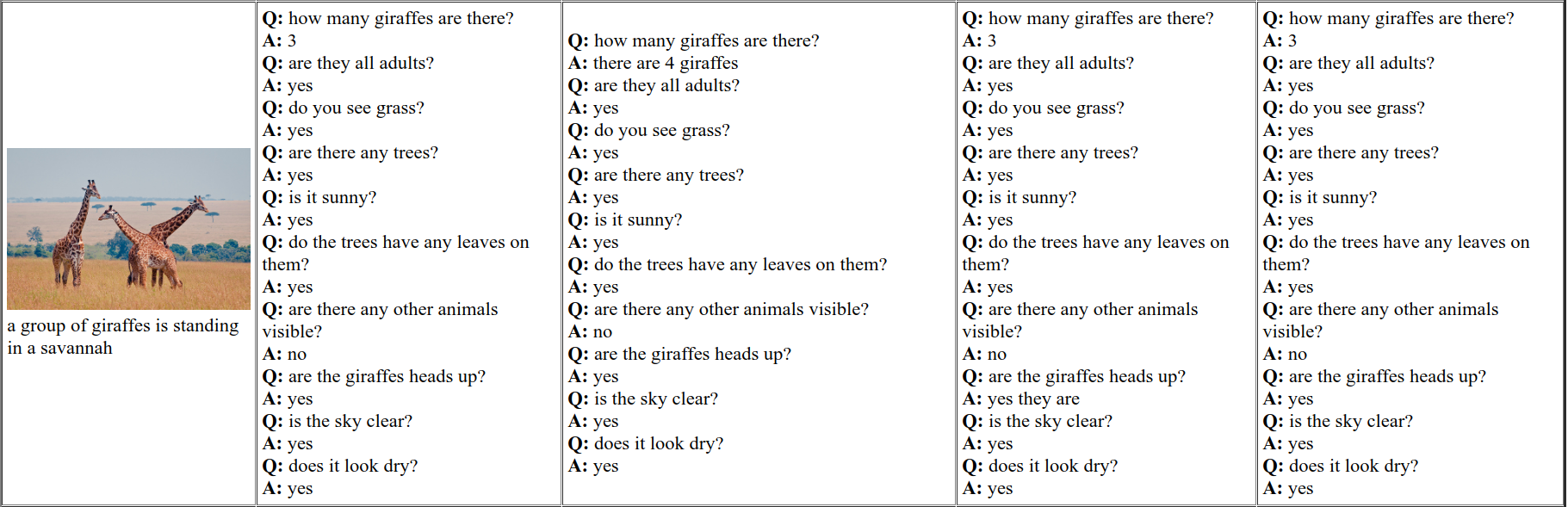}
    \end{subfigure}
    \begin{subfigure}[t]{\textwidth}
        \includegraphics[width=\textwidth]{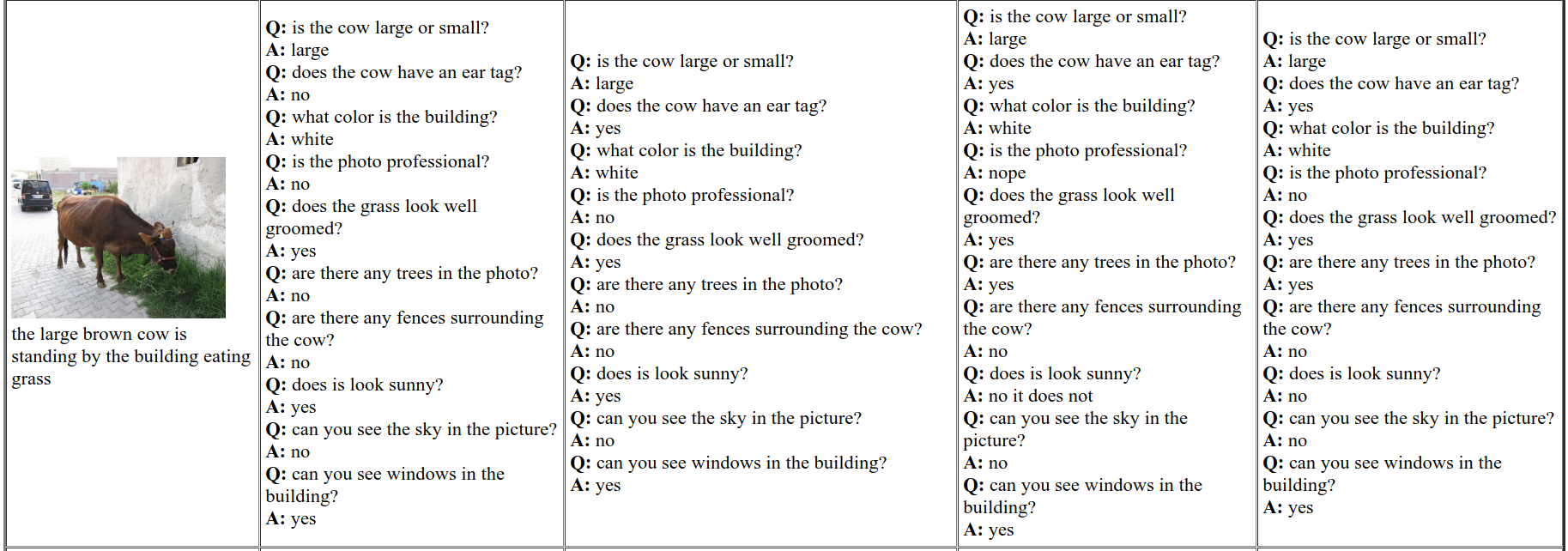}
    \end{subfigure}%
    \vspace{5pt}
    \caption{Qualitative samples for three model variants -- ViLBERT w/ CC + VQA (called `Base'), Base + CE, and Base + CE + NSP}
    \label{fig:qual1}
\end{figure*}

\textbf{Why are GT and dense annotations misaligned?}
We notice that many
questions with discrepancy between GT and dense annotations are
somewhat subjective. For~\eg, in row 1, round 7 (\figref{fig:qual1}),
Q: `what color is the chair?', the GT answer is `black' but the chair is in
shadow and it is difficult to accurately identify its color. And thus, we
expect to see variance when multiple humans are polled for the answer. Instead,
the GT answer is just one sample from the human answer distribution, not
necessarily from its peak. In general, the dense annotations seem less
wrong than GT (as they are sourced by consensus) since they are safer --
often resolving to answers like `I cannot tell' when there is uncertainty / subjectivity --
but also uninformative -- not conveying additional information~\eg `I think $3$
but they are occluded so it is hard to tell' -- since such nuanced answers are not
part of the list of answer options in VisDial~\cite{visdial}.

\textbf{Model performance on GT~\vs dense annotations}.
\tableref{tab:relevance_distribution_gt_answer} shows mean ranks of these
GT answers as predicted by three model variants -- ViLBERT w/ CC + VQA, CE, and
CE + NSP -- grouped by dense relevance scores. The `CE' model
gets worse mean ranks than `w/ CC + VQA' for all GT
answers, since it is no longer trained with these GT answers during
dense annotation finetuning. The CE model assigns low mean ranks to GT answers with
higher relevance scores ($\geq 0.8$), which translates to a high NDCG score
(\tableref{tab:val_results}). But it assigns poor mean ranks to GT answers with
relatively lower relevance scores ($\leq 0.8$), and since ${\sim}50\%$ GT answers
have relevance scores $\leq 0.8$, this hurts MRR, R@$k$, MR for the CE model
(\tableref{tab:val_results}).

Next, we consider the top-$50$ most-relevant answer options (occurring $\geq 10$ times)
as per dense annotations in VisDial v$1.0$ val (not restricting ourselves to
only GT answers).
\figref{fig:all_val_top_50_options_avg_relevance} shows the mean relevance
scores for this set, and~\figref{fig:all_val_top_50_options_mean_rank} shows
the mean ranks assigned to these answers by our models. The CE model gets better
mean ranks in this set compared to Base, leading to high NDCG.

\textbf{Qualitative examples}.
Finally, we present uniformly sampled example answer predictions on VisDial v$1.0$ val
from our models along with the ground-truth dialog sequences
in~\figref{fig:qual1} and present additional samples in the appendix.
In these examples, consistent with the Visual Dialog task definition~\cite{visdial}, at every round of dialog, the model gets the image, ground-truth human dialog history (including caption), and follow-up question as input, and predicts the answer. Specifically, the model ranks $100$ answer options. Here we show the top-$1$ prediction.

We make a few observations. 1) The Base model is surprisingly accurate,~\eg
in row 2, round 1 (\figref{fig:qual1}), Q: `can you see any people?', predicted answer: `part of
a person', in row 2, round 10, Q: `anything else interesting about the photo?',
predicted answer: `the dog is looking up at the person with his tongue out'.
2) The CE model often answers with generic responses (such as `I cannot tell'),
especially for questions involving some amount of subjectivity / uncertainty,
~\eg in row 1, round 7, Q: `what color is the chair?', predicted answer:
`I cannot tell' (the chair seems to be in shadow in the image), in row 2,
round 7, Q: `does the dog look happy?', predicted answer: `I can't tell'
(subjective question). 3) This also highlights a consequence of misalignment
between ground-truth and dense annotations. While the ground-truth
answer provides \emph{one} reasonable response for the question asked, it is
answerer-specific to quite an extent and there may be other correct
answers (annotated in the dense annotations). A negative effect of this
misalignment is that when finetuned on dense annotations (CE),
the model gets rewarded for generic answers (\eg `cannot tell'). While being
able to capture and reason about uncertainty is a desirable property
models should have, it would be more helpful if these agents
can convey more information with appropriate qualifiers (\eg `I think $3$
but they are occluded so it is hard to tell') than a blanket `I cannot tell'.
We aim to study this in future work.

\section{Implementation}
We use the BERT\textsubscript{BASE} model~\cite{devlin_naacl19}, which has $12$ layers of
Transformer blocks with each block having $12$ attention heads and a hidden state
size of $768$, for the linguistic stream. We use $6$ layers of Transformer blocks, with
each block having $8$ attention heads with a hidden state size of $1024$, for the visual stream. The co-attention layers connect the $6$ Transformer layers in the visual stream to the last $6$
Transformer layers in the linguistic stream as in~\cite{lu_neurips19}. We train on dialog sequences with atmost $256$ tokens as most sequences had atmost $256$ tokens. During inference, we truncate longer sequences by removing rounds starting from the first round (we do not remove the caption). We set all loss coefficients to $1$. All experiments use 8 GPUs (with a batch size of 128 for language-only experiments and 80 for other experiments). We use Adam~\cite{kingma_iclr15} and linearly increase learning rate from $0$ to $2\text{e}^{-5}$ over $10k$ iterations and decay to $1\text{e}^{-5}$ over $200k$ iterations.
All models were implemented in PyTorch~\cite{paszke_pytorch}.
\section{Conclusion}
\label{sec:conc}
\vspace{-5pt}

We introduce a model for Visual Dialog that enables pretraining on
large-scale image-text datasets before transferring and finetuning on VisDial.
Our model is an adaptation of ViLBERT~\cite{lu_neurips19}, and our best
single model is pretrained on BooksCorpus~\cite{zhu_iccv15}, English Wikipedia
(at the BERT stage), and on Conceptual Captions~\cite{sharma_acl18}, VQA~\cite{antol_iccv15}
(at the ViLBERT stage), before finetuning on VisDial, optionally with dense
annotations. Our model outperforms prior published results by $> 1\%$ absolute
on NDCG and MRR, achieving state-of-the-art results, and providing a simple
baseline for future `pretrain-then-transfer' approaches.

Through careful analysis of our results, we find that the recently released
dense annotations for the task do not correlate well with the original ground-truth
dialog answers, leading to a trade-off when models optimize for metrics that
take into account these dense annotations (NDCG)~\vs the original sparse annotations (MRR).
This opens up avenues for future research into better evaluation metrics.

Finally, note that our model is discriminative -- it can
pick a good answer from a list of answer options -- but cannot generate an answer.
In the future, we aim to develop robust decoding techniques for a powerful generative model.

\vspace{-10pt}
\section{Acknowledgements}
\label{sec:ack}
\vspace{-5pt}

We thank Jiasen Lu and Stefan Lee for helpful discussions.
The Georgia Tech effort is supported in part by NSF, AFRL, DARPA, ONR YIPs, ARO PECASE.
AD is supported in part by fellowships from Facebook, Adobe, and Snap Inc.
The views and conclusions contained herein are those of the authors
and should not be interpreted as necessarily representing the official policies
or endorsements, either expressed or implied, of the US Government, or any
sponsor.

\clearpage
{\small
\bibliographystyle{ieeetr}
\bibliography{main}
}
\clearpage
\section{Appendix}

\subsection{Negative Results}
To encourage the model to learn sentence level semantics, we tried a pretraining strategy which we refer to as the inconsistency loss. During pretraining, we randomly select answers at $3$ different rounds and replace them randomly with one of the $100$ answer options at those rounds. Similar to the masked language modeling loss, at each token, we predict the probability of the token being ``inconsistent'' in the dialog history and we assume that all the tokens in the randomly selected answer option are ``inconsistent''. We also only consider dialog sequences which have at least 6 rounds for creating samples. This is to make sure that the model has enough context to make accurate predictions. We hoped that this loss would encourage the model to capture sentence level semantics as the model would need to figure out which sentences fit in together in a dialog sequence.

A pitfall in this setting is that some of the $100$ answer options at each round are often similar to the GT answer or are generic responses (\eg ``yes'', ``no'', ``maybe'', \etc). Thus, there is a chance that swapping the GT answer with a randomly selected answer option might lead to a consistent dialog sequence. We instead try to create a new sample by randomly selecting a round and reordering/jumbling the answers at that round and the answers at the preceding and the following rounds. We hope that jumbling the order of answers would lead to an inconsistent dialog sequence. We call this variant ``inconsistency loss (jumbled)''.

We cannot use the batch of data used to calculate the NSP and MLM loss to calculate the inconsistency loss. We first try to use half the batch to calculate the inconsistency loss and the other half to calculate the NSP and MLM losses. We then try to calculate the inconsistency loss and the NSP and MLM losses in alternating batches. We present results for models trained with multiple variants of the inconsistency loss in the language-only setting in Table \ref{tab:inconsistency_results}.

We do not see any significant improvement by training with the inconsistency loss. We note that creating samples by reordering answers in different rounds does not improve performance. We also note that optimizing for the inconsistency loss and the NSP and MLM losses in alternating batches leads to improvements.

\begin{table}[ht!]
    \resizebox{\columnwidth}{!}{
    \begin{tabular}{@{}lcccccccc@{}}
            & \multicolumn{6}{c}{}  \\[0.01in]
          Variant & NDCG $\uparrow$ & MRR $\uparrow$ & R@$1$ $\uparrow$ & R@$5$ $\uparrow$ & R@$10$ $\uparrow$ & MR $\downarrow$ \\[0.05in]
            \toprule
    nsp + lm & $\mathbf{57.26}$ & $\mathbf{64.40}$ & $\mathbf{50.31}$ & $\mathbf{81.61}$ & $\mathbf{90.43}$ & $\mathbf{4.01}$ \\
	nsp + lm + inconsistency & $56.22$ & $64.13$ & $49.88$ & $81.35$ & $90.11$ & $4.01$ \\
	nsp + inconsistency & $55.66$ & $63.00$ & $48.63$ & $80.61$ & $89.36$ & $4.20$ \\
	nsp + inconsistency(jumbled) & $55.83$ & $63.08$ & $48.71$ & $80.47$ & $89.49$ & $4.19$ \\
	nsp + lm + inconsistency (alternating batches) & $57.17$ & $64.21$ & $50.04$ & $81.48$ & $90.02$ & $4.03$ \\
        \bottomrule
        \end{tabular}
    }
    \vspace{5pt}
    \caption{{\small Performance of language-only models on VisDial v$1.0$ val, trained with different sets of pretraining losses including the inconsistency loss.}}
    \label{tab:inconsistency_results}
\end{table}

\subsection{Additional Qualitative Samples}
\begin{figure*}[]
    \begin{subfigure}[t]{\textwidth}
        \includegraphics[width=\textwidth]{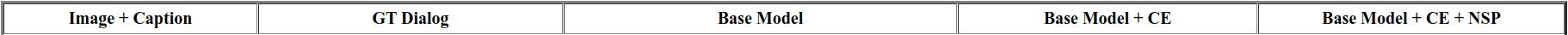}
    \end{subfigure}
    \begin{subfigure}[t]{\textwidth}
        \includegraphics[width=\textwidth]{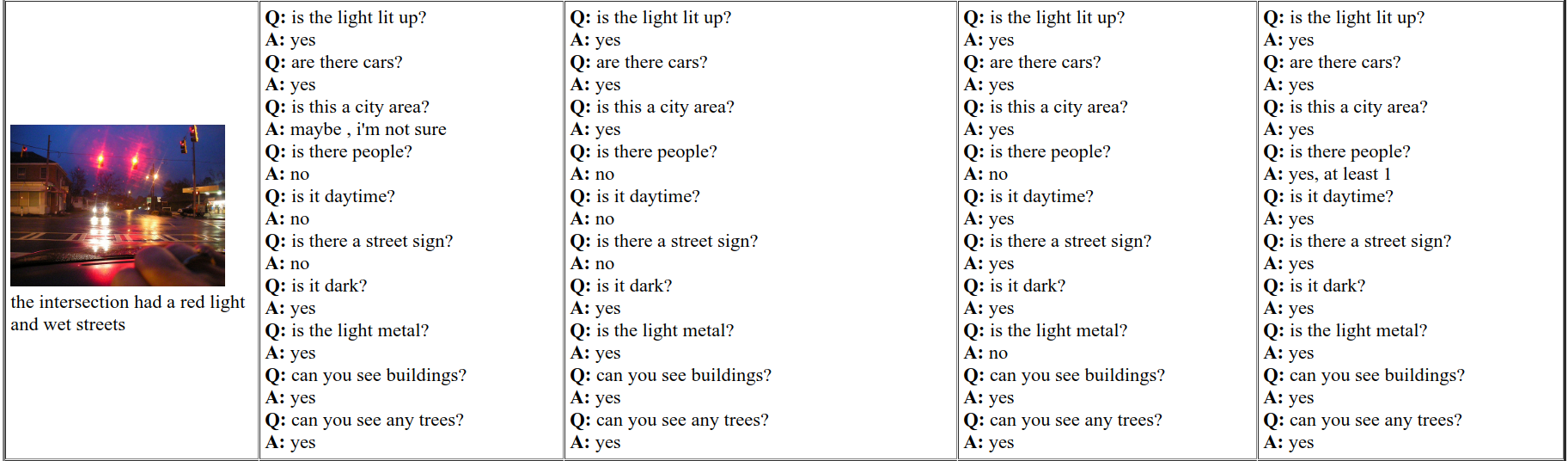}
    \end{subfigure}
    \begin{subfigure}[t]{\textwidth}
        \includegraphics[width=\textwidth]{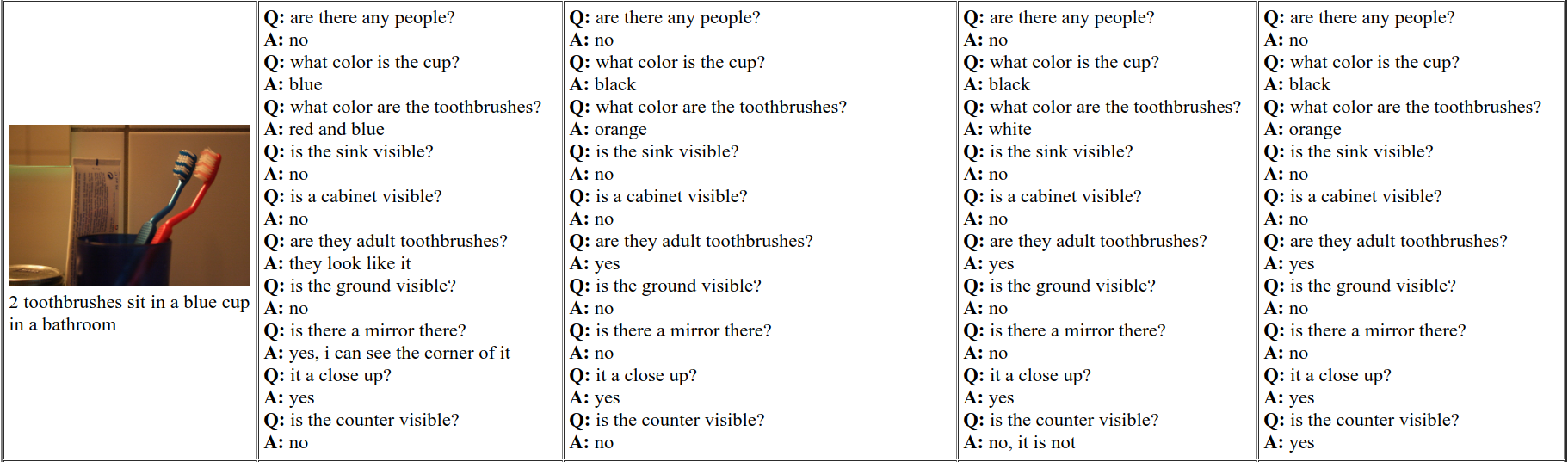}
    \end{subfigure}
    \begin{subfigure}[t]{\textwidth}
        \includegraphics[width=\textwidth]{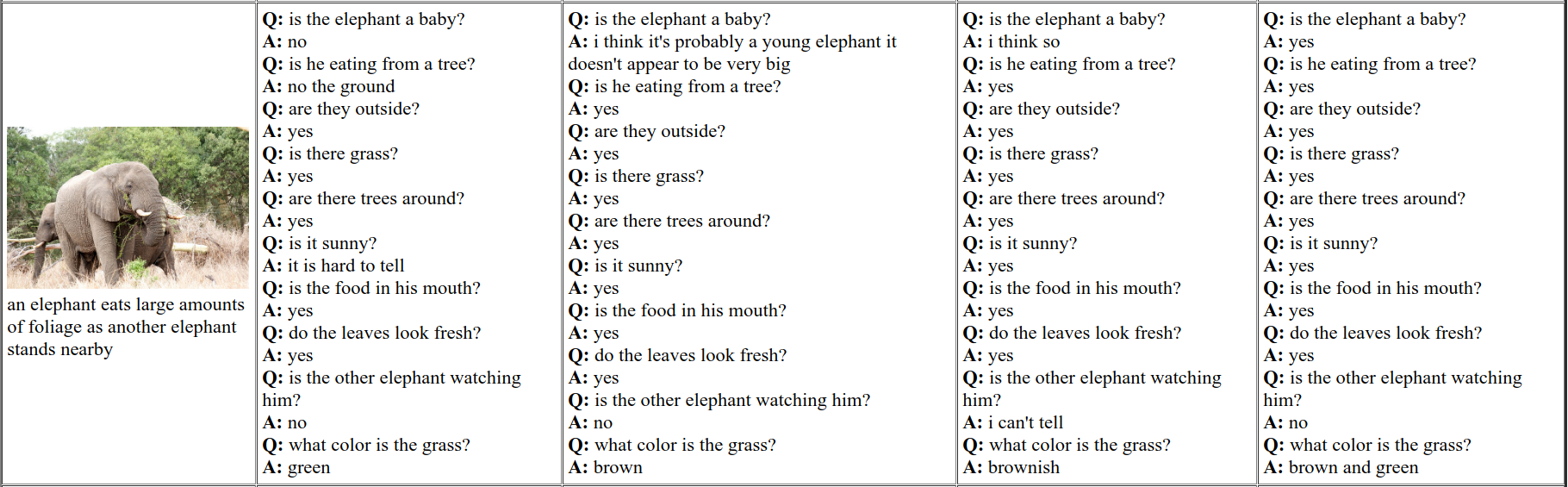}
    \end{subfigure}
    \begin{subfigure}[t]{\textwidth}
        \includegraphics[width=\textwidth]{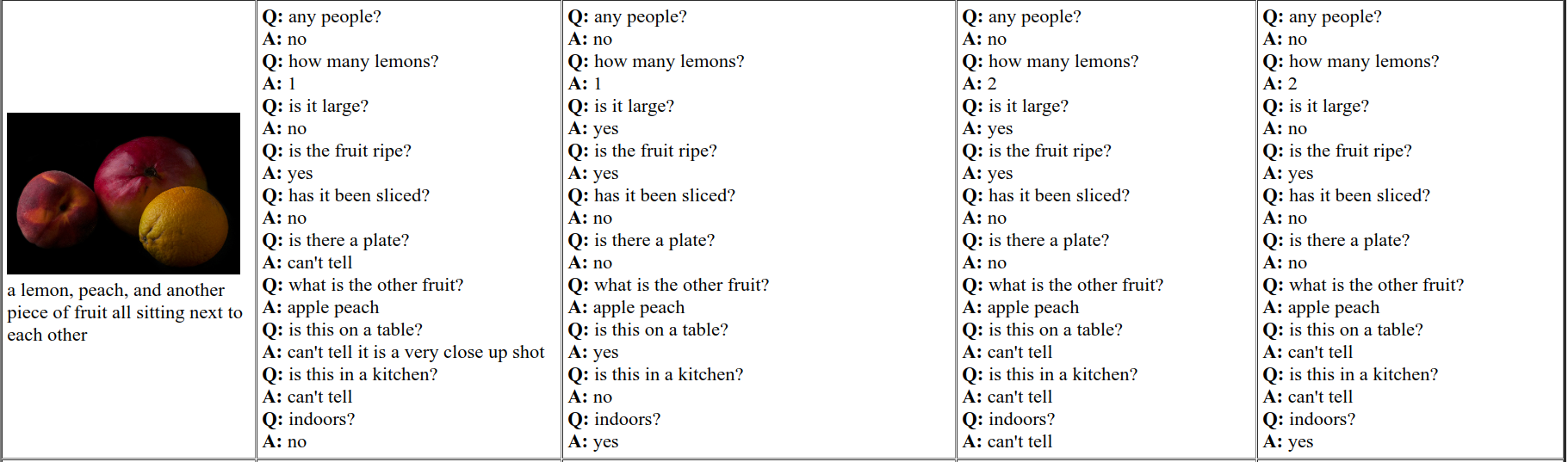}
    \end{subfigure}%
    \vspace{5pt}
    \caption{Qualitative samples for three model variants -- ViLBERT w/ CC + VQA (called `Base'), Base + CE, and Base + CE + NSP.}
    \vspace{-20pt}
    \label{fig:qual2}
\end{figure*}

\begin{figure*}[]
    \centering
    \begin{subfigure}[]{0.92\textwidth}
        \includegraphics[width=\textwidth]{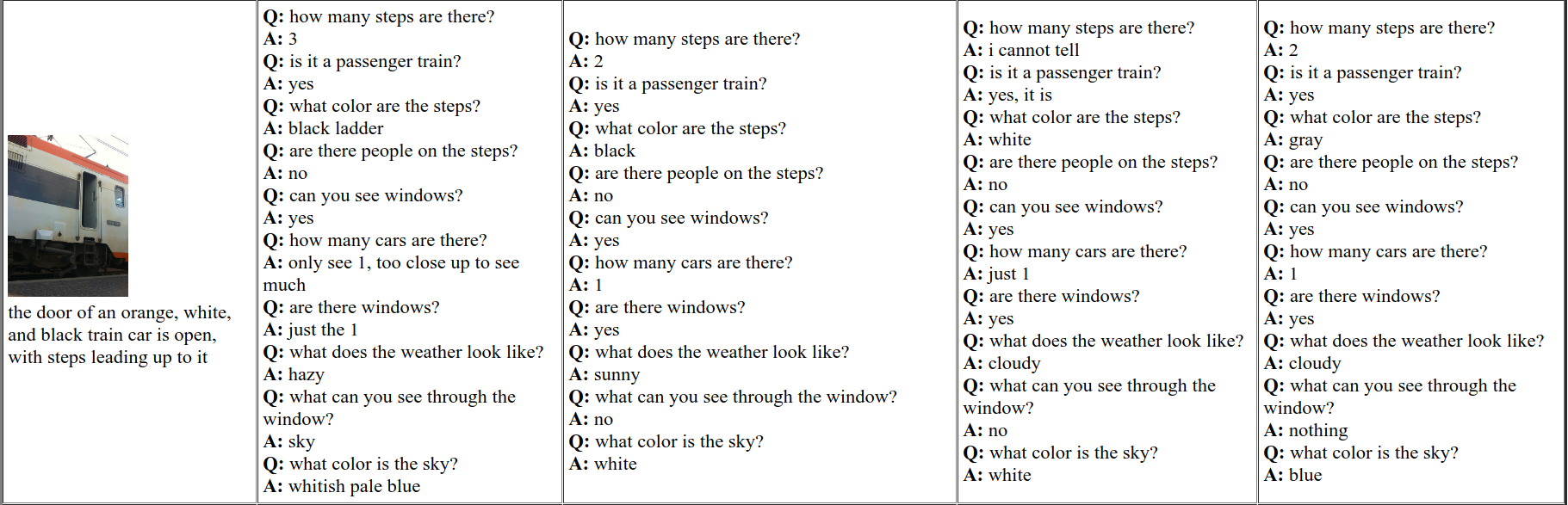}
    \end{subfigure}
    \begin{subfigure}[]{0.92\textwidth}
        \includegraphics[width=\textwidth]{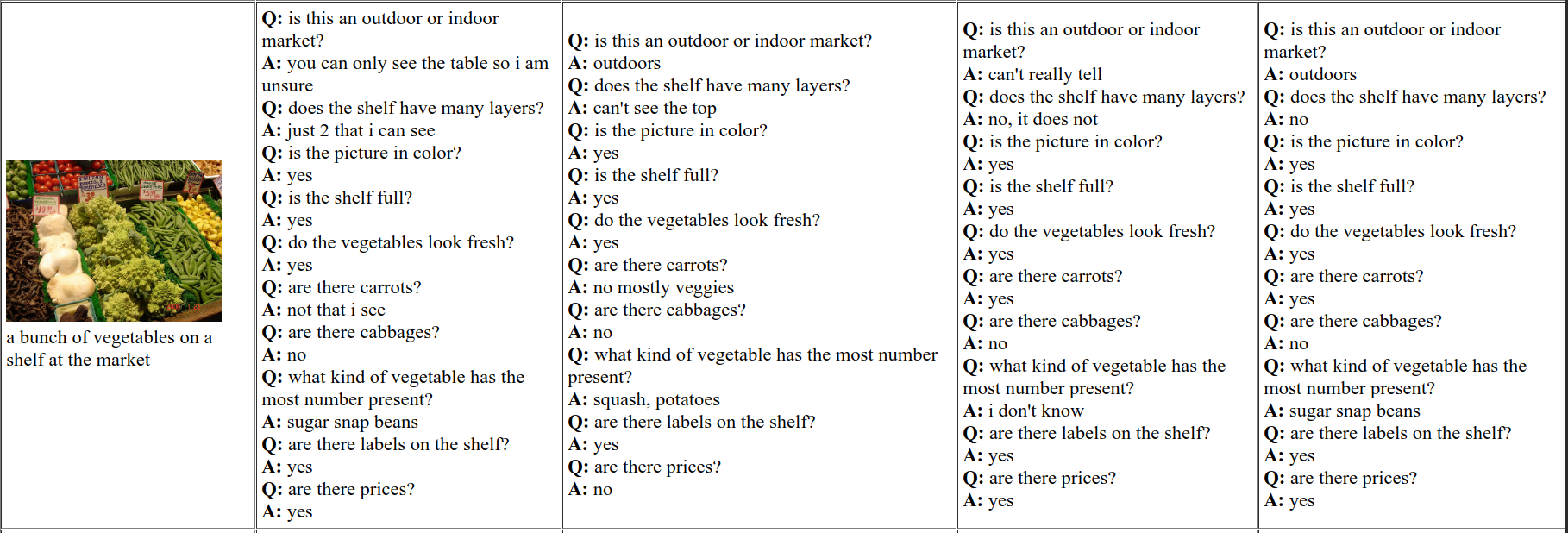}
    \end{subfigure}
    \begin{subfigure}[]{0.92\textwidth}
        \includegraphics[width=\textwidth]{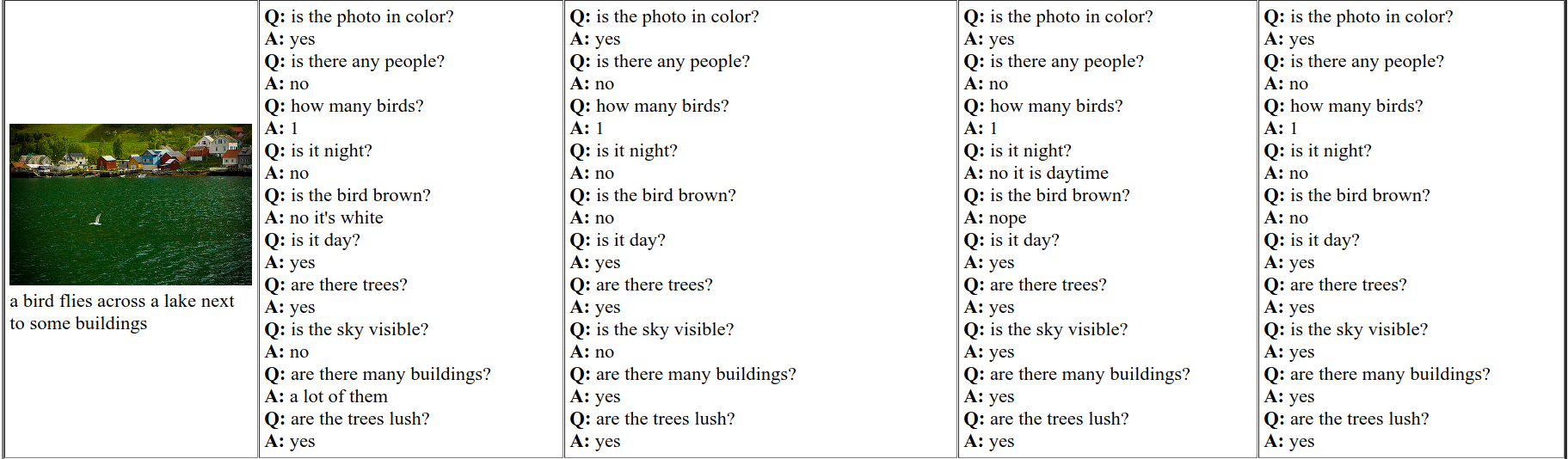}
    \end{subfigure}
    \begin{subfigure}[]{0.92\textwidth}
        \includegraphics[width=\textwidth]{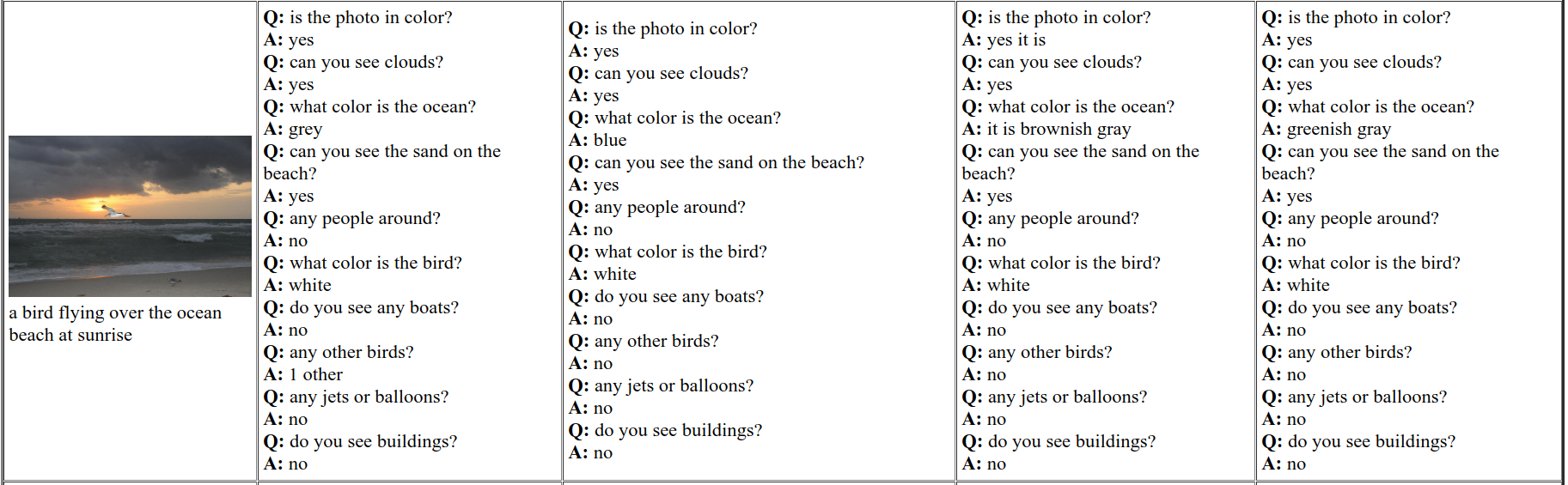}
    \end{subfigure}%
    \vspace{5pt}
    \caption{Qualitative samples for three model variants -- ViLBERT w/ CC + VQA (called `Base'), Base + CE, and Base + CE + NSP.}
    \label{fig:qual3}
\end{figure*}

We present more qualitative samples in  \figref{fig:qual2} and \figref{fig:qual3}.

\end{document}